\documentclass{article}

    \PassOptionsToPackage{numbers, compress}{natbib}

\usepackage[preprint]{neurips_2024}

\usepackage[utf8]{inputenc} 
\usepackage[T1]{fontenc}    
\usepackage{hyperref}       
\usepackage{url}            
\usepackage{booktabs}       
\usepackage{amsfonts}       
\usepackage{nicefrac}       
\usepackage{microtype}      
\usepackage{xcolor}         
\usepackage{algorithm}
\usepackage{algorithmic}
\usepackage{amssymb}
\usepackage{wrapfig}
\usepackage{bbding}         
\usepackage{colortbl} 
\hypersetup{
    colorlinks=true,
    linkcolor=red,
    filecolor=black,
    citecolor=green,
    urlcolor=red,
}

\title{\textit{Holmes-VAD}: Towards Unbiased and Explainable\\ Video Anomaly Detection via Multi-modal LLM}

\author{%
  Huaxin Zhang$^{1,3}$,
  Xiaohao Xu$^{2}$,
  Xiang Wang$^{1}$,
  Jialong Zuo$^{1}$,
  Chuchu Han$^{1,3}$,\\
  \textbf{Xiaonan Huang$^{2}$,
  Changxin Gao$^{1}$,
  Yuehuan Wang$^{1}$,
  Nong Sang$^{1}\textsuperscript{\Envelope}$}
  \\
  $^1$Key Laboratory of Image Processing and Intelligent Control,\\
  School of Artificial Intelligence and Automation,\\
Huazhong University of Science and Technology \\
  $^2$University of Michigan, Ann Arbor \quad
  $^3$Baidu Inc. \quad
  {\Envelope~Corresponding Author} \\
}

\usepackage{multirow} 
\usepackage{multicol} 
\usepackage{graphicx} 
\usepackage{caption}
\usepackage{subcaption}

\usepackage{colortbl}  
\usepackage{xcolor}
\usepackage{pifont}
\usepackage{array}   
\usepackage{amsmath}
\newcommand{\ie}{\textit{i}.\textit{e}.}
\newcommand{\eg}{\textit{e}.\textit{g}.}
\newcommand{\etal}{\textit{et}~\textit{al}.} 
\newcommand{\True}{\textcolor{green}{\ding{51}}}
\newcommand{\False}{\textcolor{red}{\ding{55}}}
\definecolor{Gray}{gray}{0.94}

\begin{document}

\maketitle
\begin{abstract}
Towards open-ended Video Anomaly Detection (VAD), existing methods often exhibit biased detection when faced with challenging or unseen events and lack interpretability. To address these drawbacks, we propose Holmes-VAD, a novel framework that leverages precise temporal supervision and rich multimodal instructions to enable accurate anomaly localization and comprehensive explanations.
Firstly, towards unbiased and explainable VAD system, we construct the first large-scale multimodal VAD instruction-tuning benchmark, \textit{i.e.}, \textit{VAD-Instruct50k}. This dataset is created using a carefully designed semi-automatic labeling paradigm. Efficient single-frame annotations are applied to the collected untrimmed videos, which are then synthesized into high-quality analyses of both abnormal and normal video clips using a robust off-the-shelf video captioner and a large language model (LLM).
Building upon the \textit{VAD-Instruct50k} dataset, we develop a customized solution for interpretable video anomaly detection. We train a lightweight temporal sampler to select frames with high anomaly response and fine-tune a multimodal large language model (LLM) to generate explanatory content.
Extensive experimental results validate the generality and interpretability of the proposed \textit{Holmes-VAD}, establishing it as a novel interpretable technique for real-world video anomaly analysis.
To support the community, our benchmark and model will be publicly available at \url{https://holmesvad.github.io/}.
\end{abstract}

\section{Introduction}
\label{sec:introduction}
Video Anomaly Detection (VAD)~\cite{ConvAE} aims to identify abnormal events in videos, which has been extensively researched in recent years due to its considerable application value in public safety~\cite{ucf} and video content understanding~\cite{xdviolence}.
Current VAD approaches can be broadly classified into three categories according to the annotation type of the training data, \ie, unsupervised, weakly-supervised and fully-supervised.
Unsupervised methods~\citep{ConvAE,rnn,framepred,gong2019memorizing,GODs,yang2023video} train solely on normal videos (one-class) or unlabeled normal/abnormal videos, while weakly supervised methods~\citep{ucf,GCN,rtfm,MSL,S3R,URDMU,UMIL} train on normal/abnormal videos with video-level labels. Fully-supervised methods~\citep{liu2019exploring,landi2019anomaly} are less studied due to the high cost of precise frame-by-frame annotations.
Recently, inspired by the strong representation of multi-modal large language models (MLLMs) pretrained on massive data~\citep{touvron2023llama,chiang2023vicuna,jiang2023mistral,yang2023mm,lin2023mm,zhu2023minigpt,liu2024visual,zhang2023llama,ye2023mplug,dai2024instructblip,bai2023qwen,wang2023cogvlm} and their impressive advancements in many downstream visual tasks~\citep{gu2024anomalygpt,wang2023clip,wang2023few}, many efforts~\citep{PromptEnhanced,cliptsa,vadclip,yang2024text,wu2023open,zanella2024harnessing} start to integrate the multi-modal knowledge into VAD systems, which enables more 
{precise}
anomaly detection.
Despite significant progress, existing VAD models still face the following primary challenges:
\begin{itemize}
    \item{\textbf{Biased anomaly space}: 
    Due to the lack of reliable frame-level abnormal supervision, unsupervised methods fail to reconstruct or predict unseen normal data, while weakly supervised methods also struggle to select trustworthy snippets for training under the video-level supervision. Consequently, the learned anomaly space of these methods develop a prevalent bias toward unseen or easily-confused normality, remain \textbf{"when does the anomaly happen"} still facing challenges.
    Although there are some fully supervised studies~\citep{liu2019exploring,landi2019anomaly}, the number of annotated videos is very small due to the inefficiency of the annotation process, resulting in a lack of scalability.
    }
    \item{\textbf{Lack of explainability}: 
    {Existing}
    video anomaly detection approaches do not offer transparent explanations and reasoning for detected anomalies, \ie, \textbf{"what is the anomaly"}, and \textbf{"why is it considered anomalous"}. This opacity restricts human comprehension and engagement with the system.
    }
\end{itemize}
\begin{figure*}[t]
\centering
\includegraphics[width=\textwidth]{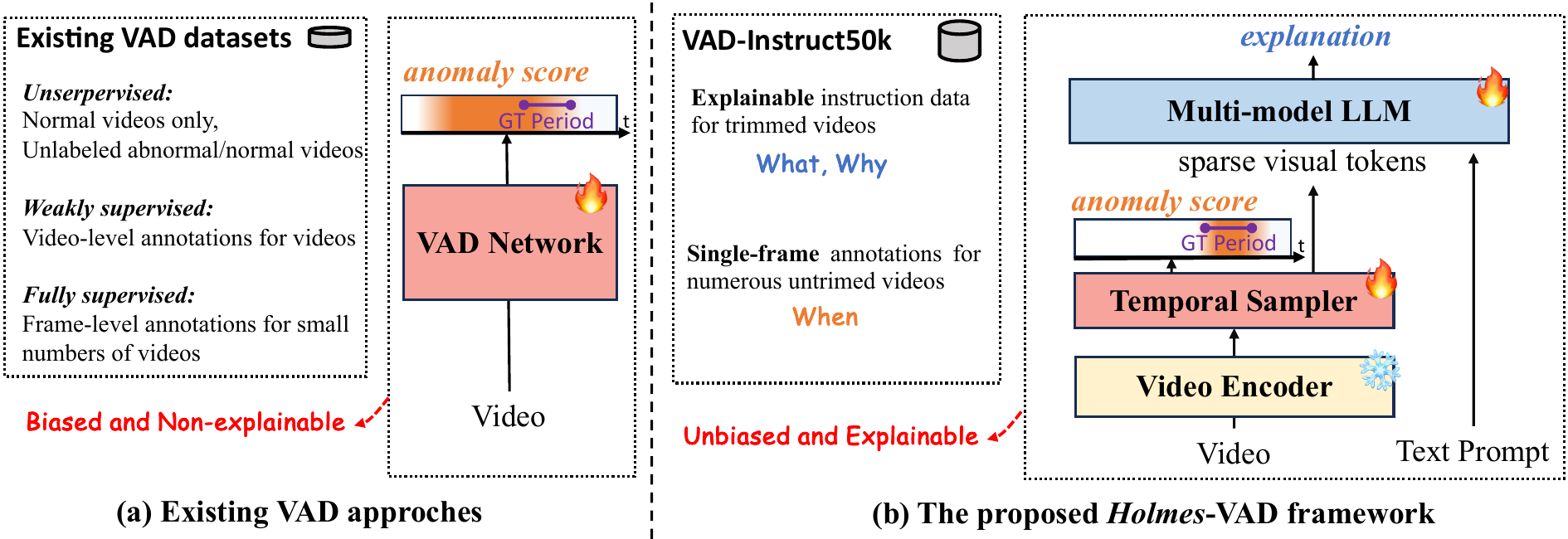}
\caption{
\textbf{Towards unbiased and explainable VAD}.
In contrast to prevailing VAD approaches (\textbf{a}) that primarily concentrate on identifying anomalies, our method (\textbf{b}) facilitates not only \textbf{unbiased} (\textit{i.e.,} less false alarms toward 
 easily cofused or unseen normality) predictions of anomaly scores but also \textbf{explanation} of detected anomalies, through constructing a large scale VAD dataset with single-frame annotations for untrimmed videos and explanable instruction data for trimmed videos.
}
\vspace{-4mm}
\label{fig:motivation}
\end{figure*}

Drawing from the above analysis, our insight is that a strong AI-powered anomaly detection system requires not only identifying deviations, but also providing insightful explanations, mirroring the deductive reasoning like the detective \textit{\textbf{Sherlock Holmes}}.
To this end, we present \textit{Holmes-VAD}, an unbiased and explainable VAD framework based on MLLMs (see Fig.\ref{fig:motivation}).
More specifically, to tackle the first issue,
we propose a more label-friendly single-frame supervision (one-click for each abnormal event)~\citep{sf,lacp,hrpro,cui2022video,li2023d3g} in the domain of video anomaly detection instead of the prohibitive frame-by-frame annotation.
Following this labeling paradigm, we manually make single-frame annotations for the exsiting two largest VAD datasets, \eg, UCF-Crime~\cite{ucf} and XD-Violence~\cite{xdviolence}.
To address the second problem of lacking explainability, we construct a large amount of anomaly-awared instruction conversation data for the finetuning of Multimodal LLM.
We leverage the single-frame annotated videos and exsiting off-the-shell large foundation model to build an efficient semi-automated data engine.
This data engine can be divided into three main steps: 1) \textbf{Data Collection}: gathering video data, primarily from open-source datasets. 2) \textbf{Annotation Enhancement}: generate reliable video event clips around the single-frame annotated frames and give textual descriptions to them through human effort or foundation models. 3) \textbf{Instruction Construction}: utilizing powerful LLM with open-world knowledge to generate explanable analysis in the context of the enhanced video annotations.
Subsequently, the obtained analysis is filted manually and structured into conversational format.
After on the above steps, a new benchmark containing single-frame temporal annotations and explanatory text descriptions is constructed, and we name the final obtained dataset as \textbf{VAD-Intruct50k}.
Built upon the proposed VAD-Intruct50k, we develop a customized solution for interpretable video anomaly detection, which has three key components, \ie, Video Encoder, Temporal Sampler and Multi-modal LLM.
The Video Encoder and Multi-modal LLM are used to encode the input video and generate text response to the input text prompt, respectively.
Additionally, the Temporal Sampler is used to predict the abnormal scores of video frames and sample high-responsive parts as the input for Multi-modal LLM, which is lightweight and enables effient inference.
Specially, these three components can be replaced by any other Video-MLLMs or VAD-Networks.
Our primary focus is on how to construct a supervised multi-modal dataset to train these components.
Extensive experiments demonstrate that our \textit{Holmes}-VAD achieve outstanding performance in video anomaly detection and can provide detailed explanations for the detected abnormal events.

To summarize, our major contributions are as follows: 
\begin{itemize}
\item We propose \textbf{\textit{Holmes-VAD}},  a video anomaly detection system that is capable of identifying anomalies and providing insightful explainations across even hour-long videos.\

\item To bridge the dataset gap toward an unbiased and explanable VAD system, we introduce \textbf{VAD-Intruct50k}, a large-scale multimodal video anomaly detection datasets, including single-frame annotations for untrimmed videos, and a large amount of instruction conversation data for trimmed abnormal/normal video clips.

\item Extensive quantitative and qualitative experiments demonstrate that the proposed \textbf{\textit{Holmes-VAD}} achieves superior performance and  interpretability over recent state-of-the-art methods.
\end{itemize}

\section{Related Works}

\noindent\textbf{Video Anomaly Detection.}
This task aims to temporally detect abnormal frames in a long untrimmed video~\citep{adam2008robust,mehran2009abnormal,li2013anomaly,Luetal,GODs,ConvAE}.
The early VAD attempts are based on hand-crafted features~\citep{adam2008robust,kim2009observe,zhao2011online,mehran2009abnormal,Luetal,li2013anomaly}.
Recently, deep learning approaches~\citep{ConvAE,yang2023video,UMIL} have become dominant in Video Anomaly Detection (VAD), broadly classified into unsupervised, weakly-supervised, and fully-supervised methods. Unsupervised methods train only on normal videos to learn normal patterns and are often designed as reconstruction-based~\citep{ConvAE,xu2017detecting,gong2019memorizing,yang2023video}, prediction-based~\citep{framepred}, or a combination~\citep{liu2021hybrid}. Some methods~\citep{zaheer2022generative,thakare2023dyannet,tur2023exploring} also explore a fully unsupervised setting, including both unlabeled normal and abnormal videos in the training set. Weakly-supervised methods~\citep{ucf,GCN,mist,xdviolence,rtfm,MSL,S3R,URDMU,UMIL,zhang2024glancevad} use both normal and abnormal videos with video-level annotations. Fully-supervised methods~\citep{liu2019exploring,landi2019anomaly} are less common due to the high cost of precise frame-level annotations.

\noindent\textbf{Multi-modal Large Language Model.}
The universal and powerful conversational capabilities of ChatGPT~\citep{achiam2023gpt} have inspired the entire AI community. 
This has prompted the emergence of the open-source Large Language Models (LMMs), such as LLaMA~\citep{touvron2023llama}, Vicuna~\citep{chiang2023vicuna}, and Mistral~\citep{jiang2023mistral}, based on autoregressive models~\citep{vaswani2017attention}, they are pretrained and instruction tuned via large amounts of text tokens, thus posses universal and powerful text generation capabilities.
Recently, Multi-modal LLMs~\citep{yang2023mm,lin2023mm,zhu2023minigpt,liu2024visual,liu2023improved,zhang2023llama,ye2023mplug,dai2024instructblip,bai2023qwen,wang2023cogvlm} empower LLMs with  visual understanding capabilities.
Additionally,  MLLMs for videos (\textit{e.g.}, VideoChat~\citep{li2023videochat}, Video-LLaMA~\citep{zhang2023video}, and Video-LLaVA~\citep{lin2023video}) pave the way for  multi-modal temporal understanding.

\noindent\textbf{Multi-modal Video Anomaly Detection.}
Large-scale visual-language pretrained models such as CLIP~\citep{radford2021learning} serve as a bridge between visual and textual modalities.
Some recent works~\citep{PromptEnhanced,cliptsa,vadclip,yang2024text} in the realm of video anomaly detection
have leveraged textual information as prompts to enhance the model's anomaly representation.
Based on this, ~\cite{wu2023open} firstly proposed the open vocabulary VAD task.
Furthermore, ~\cite{zanella2024harnessing} extracted captions from video frames using a caption model and designed prompts for LLMs to provide anomaly scores.
However, these approaches primarily focus on generating anomaly scores and lack fine-tuning on large-scale domain-specific instruction datasets, resulting in their performance being highly dependent on the base LLMs.

\section{VAD-Instruct50k Benchmark}
\label{sec:proposed_dataset}
\begin{figure*}[t]
\centering
\includegraphics[width=\textwidth]{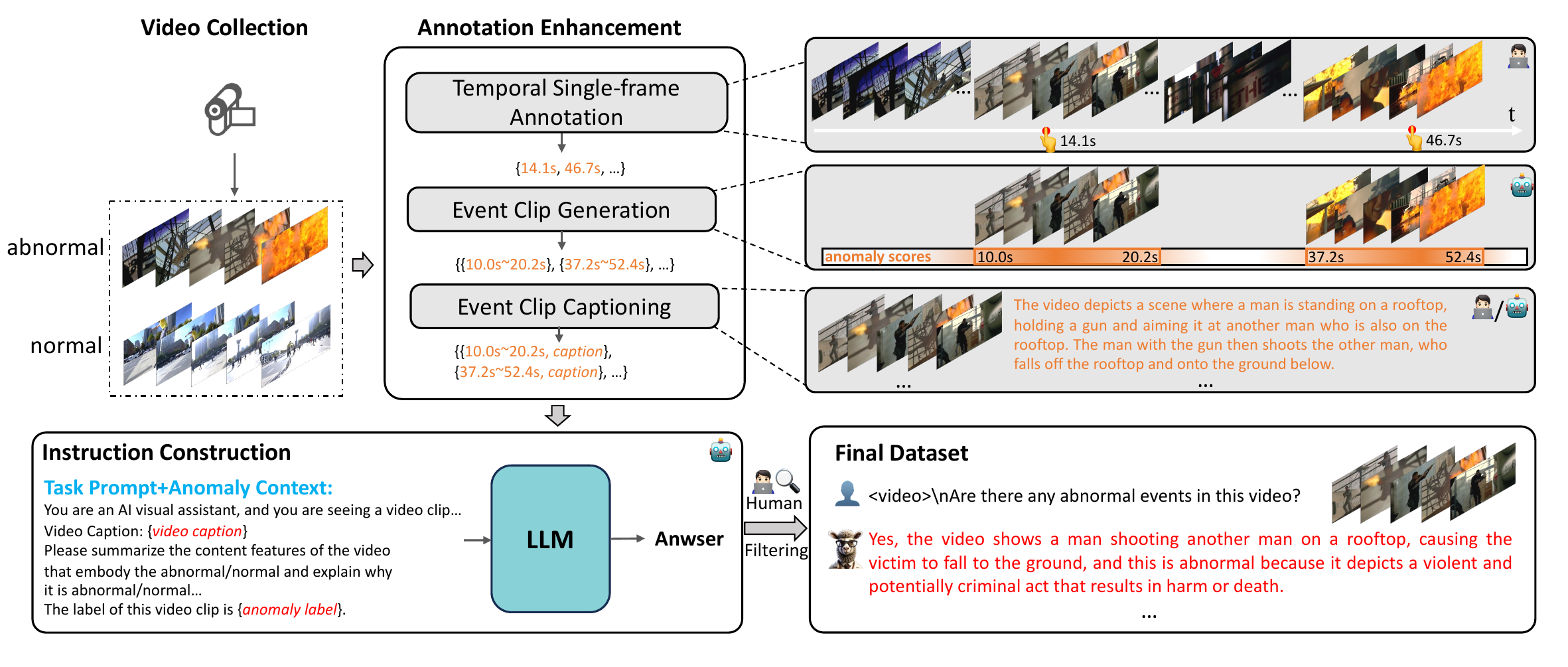}
\vspace{-4mm}
\caption{Data engine for the proposed \textit{VAD-Instruct50k}.
We collect numerous abnormal/normal videos from exsiting datasets, following by a series of annotation enhancement including temporal single-frame annotation, event clips generation and event clips captioning.
Then we construct the instruction data by prompting the powerful LLM with the enhanced annotation.
Throughout the pipeline, manual work and large fundation models coordinated with each other to ensure efficiency and quality in construction.
}
\label{fig:data_engine}
\end{figure*}

In this section, we will illustrate the process of VAD-Instruct50k dataset generation.
Firstly, the data collection process of VAD-Instruct50k will be presented.
Subsequently, we will elaborate on how to enhance the annotations of the collected videos.
Finally, the generation process of the instruction conversation data will be introduced.
The overall pipeline of the data engine is shown in Fig.~\ref{fig:data_engine}.

\subsection{Data Collection}
\label{subsection:data_collection}
We first collect videos from the training sets of the two largest weakly-supervised VAD datasets, UCF-Crime~\cite{ucf} and XD-Violence~\cite{xdviolence}, because their video quantity far exceeds that of other existing datasets~\citep{Ped2,Avenue,ShanghaiTech}, and their video-level annotations provide a solid foundation for further data processing.
After filtering out some low-quality videos via human inspection, we collected a total of 5547 untrimmed videos, include 810/800 abnormal/normal videos from UCF-Crime and 1905/2032 abnormal/normal videos from XD-Violence.

\subsection{Annotation Enhancement}
\label{subsection:anno_enhancement}
The collected videos from UCF-Crime~\cite{ucf} and XD-Violence~\cite{xdviolence} only offers video-level anomaly labels, which denotes whether the video includes anomalies. Going beyond these coarse annotations, we purify these annotations to enable more discriminative anomaly detection model training.

\textbf{Temporal single-frame annotation.}
We adopt an efficient temporal annotation method involving sparse single-frame annotation for the collected abnormal videos, inspired by \citep{mettes2016spot,sf,hrpro,lacp,li2023d3g,zhang2024glancevad} that use this approach to balance model performance and annotation cost.
Specifically, we annotate only one frame for each abnormal event in the video\footnote{More details about the annotation process are illustrated in Sec.~\ref{supp:single_frame} of the Appendix.}.
Through this process, we collect an average of 2.35 single-frame annotations per video.

\textbf{Event clip generation.}
Based on the single-frame annotation, we design a reliable pseudo frame-level label generation method and leverage it to train a VAD network $\phi_s$ \footnote{See Sec.~\ref{supp:temporal_sampler} of the Appendix for more details about the network.}.
For each abnormal video with single-frame annotations $\mathcal{G}=\{g_i\}^{N_g}$ and its anomaly score estimated by the trained VAD network, we generate multiple anomaly event proposals around the annotated frame. Formally, each proposal is represented via a starting and ending timestamp, \textit{i.e.}, $s$ and $e$.
For each normal video, we randomly extract several normal event proposals. After this process, we collect all trimmed event clips with anomaly labels: $\mathcal{E}=\{s_i,e_i,y_i\}^{N_e}$, where $y_i$ is set to the anomaly class of the video (\eg, \textit{Explosion}) if the event clip is from an abnormal video, otherwise, it is set to \textit{Normal}.

\textbf{Event clip captioning.}
To fully extract semantic information from the event clips, we utilize a video-based multimodal large language model (MLLM) ~\cite{lin2023video} to generate detailed captions for each event clip. We also include the SurveillanceVision dataset ~\citep{yuan2023surveillance}, which provides manually annotated detailed fine-grained event descriptions for video clips from UCF-Crime ~\citep{ucf}. After combining these resources, we obtain all event clips with corresponding captions $c$ and anomaly labels: $\mathcal{E}=\{s_i,e_i,y_i,c_i\}_i^{N_e}$.

\subsection{Instruction-tuning Data Construction}
\label{subsection:instruction_build}
The process of annotation enhancement effectively fills the gap of insufficient information in the original video-level annotation.
However, there is still a lack of anomaly-awared explanation for these event clips, \ie, what is the anomaly and why.
To address this issue, we utilize the powerful LLM with sufficient open-world knowledge for further instruction dataset construction.
Technically, for each event clip in $\mathcal{E}$,  we design a task prompts $P_t$ combined with the referenceable anomaly context, \ie, the abnormal label $y_i$ and the detail caption $c_i$.
Then we input the combined prompt into the LLM $\mathcal{M}$ to make a judgment on anomalies in the video clip and providing an explanation. The generated response is paired with a corresponding anomaly-awared quesion $P_d$, result in an instruction item:
\begin{equation}
    \mathcal{I}_i = \{\text{"user"}:P_d, \text{"assistant"}:\mathcal{M}(P_t, y_i, c_i)\}
\end{equation}
We use Llama3-Instruct-70B~\citep{llama3modelcard} as $\mathcal{M}$  here because of its open-source availability and comparable performance to GPT4.
We design multiple $P_d$ to ensure the diversity of the instruction data, a typical prompts of $P_d$ is: "<video>\textbackslash n Are there any unexpected or unusual events in the video clip?".

\section{Holmes-VAD}
\label{sec:proposed_model}
\begin{figure*}[t]
\centering
\includegraphics[width=\textwidth]{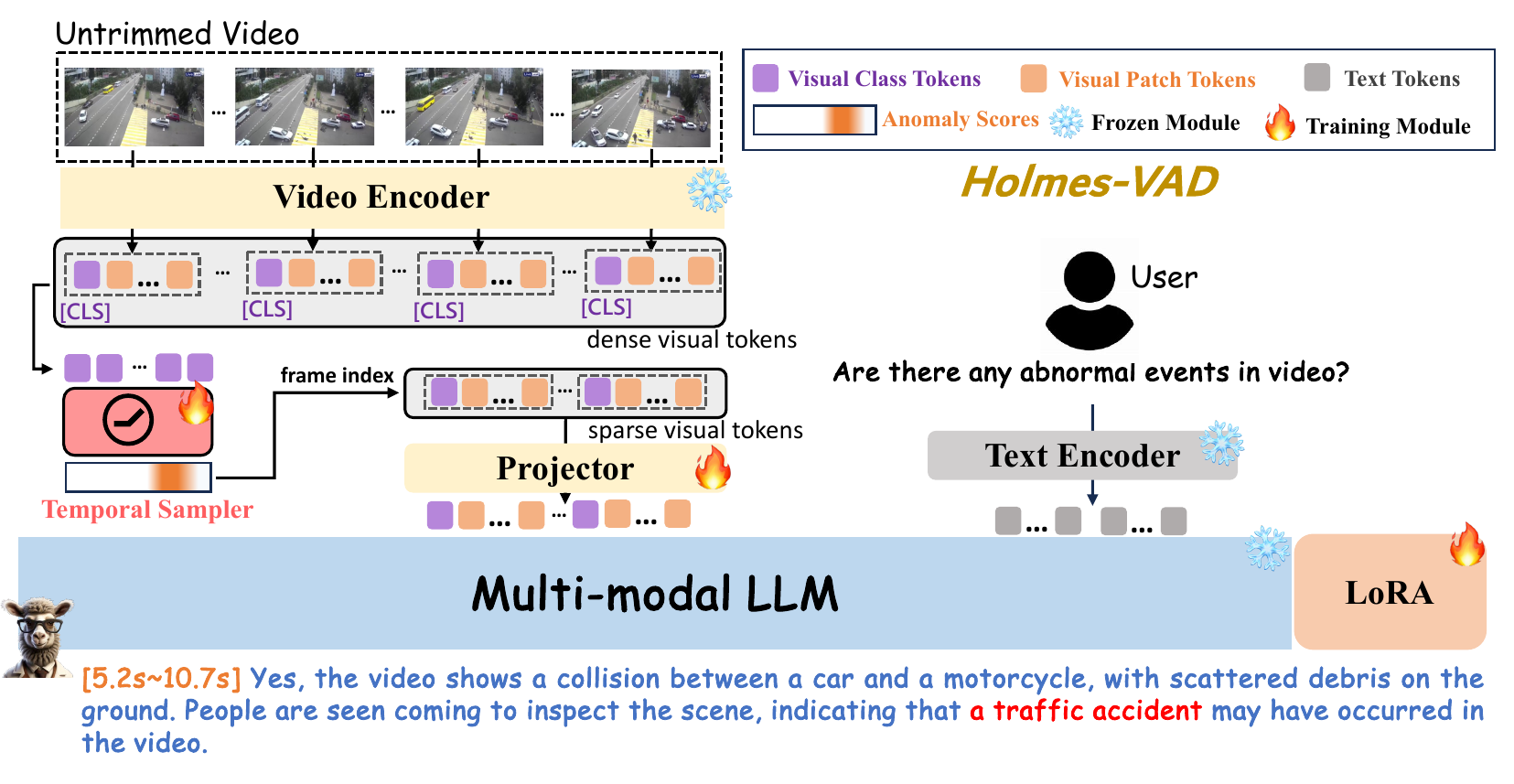}
\vspace{-4mm}
\caption{
Overview of \textit{Holmes-VAD}. 
{
\textit{Holmes-VAD} takes untrimmed video and user prompt as inputs, and takes the anomaly scores and explanation for detected anomalies outputs.
The Temporal Sampler takes class tokens of frames as input and estimates the anomaly scores, and the dense visual tokens are resampled accroding to their anomaly scores before entering the projector.
}
}
\label{fig:method_overview}
\end{figure*}

Utilizing the proposed VAD-Intruct50k dataset for training, we develop a customised solution for interpretable video anomaly detection, namely Holmes-VAD, 
which has three key components, Video Encoder, Temporal Sampler and Multi-modal LLM with tunable LoRA~\cite{hu2021lora} modules (see Fig.~\ref{fig:method_overview}).

\subsection{Model Architecture}
\textbf{Visual Encoder.} We utilize the frozen video encoder in LanguageBind~\citep{zhu2023languagebind} following~\citep{lin2023video}.
It inherits the ViT-L/14 structure from CLIP~\citep{radford2021learning}, we refer to it as $\phi_v$.
Different from the orginal ViT, it models the temporal relationship between frames through additional self-attention layer in the temporal dimension.
Give a video frame sequence ${V \in {\mathbb R}^{N \times H \times W \times C}}$, the output features of each frame can be denotes as follow:
\begin{equation}
    F^{d}_i=\{f_i^{cls}, f_i^1, f_i^2,...,f_i^{N_p}\} = \phi_v(V_i) \qquad i \in \{1,2,...,N\}
\end{equation}
where $f_i^{cls}$ indicates the class token feature of $i$-th video frame,  $f_i^{j}$ ($j \in \{1,2,...,N_p\}$) denotes the visual embedding of each patch, and $N_p$ reperesents the number of patches of each frame.

\textbf{Temporal Sampler.}
Due to the excessive computational burden caused by numerous visual tokens in video, past video-based MLLM approaches~\citep{li2023videochat,zhang2023video,lin2023video} have resorted to uniform temporal frame sampling of videos, \eg, 8 frames.
This method is clearly unsuitable for long videos in video anomaly detection task, as it increases the probability of ignoring key information.
~\citep{zanella2024harnessing} conduct dense anomaly detection via MLLM in a frame-by-frame mode, which also inevitably leads to a large amount of redundant computation.
To address this issue, we first input the dense video frames into the visual encoder, then we introduce the trained VAD network in \ref{subsection:anno_enhancement} here, which receives the cls token of the video frames $f_1^{cls},f_2^{cls},...,f_N^{cls}$ and outputs anomaly scores $s_1, s_2,...,s_N$:
\begin{equation}
    \{s_1, s_2,...,s_N\} = \phi_s(\{f_1^{cls},f_2^{cls},...,f_N^{cls}\})
\end{equation}
where $\phi_s$ denotes the trained VAD network.

Then, we sample the video tokens according to the anomaly scores. Specifically, only the tokens $f_k$ from frames with corresponding anomaly score $s_k$ above a set threshold $\theta$ are then fed into the subsequent network:
\begin{equation}
F^{s} = \{ f_k \in F^{d} \mid  s_k > \theta \} 
\end{equation}
where $F^{s}$ denotes the sampled sparse visual tokens from the original dense visual tokens $F^d$. In this way, the model can generate anomaly-awared response to long untrimmed video.

\textbf{Projector and LLM.}
To enable the LLM to understand the features output by the visual encoder, a projector ${\phi_{proj}}$ composed of two layers of MLPs is designed between them, after this, the feature dimention is aligned with the input dimension of LLM. 
We utilize Vicuna~\citep{chiang2023vicuna} as our LLM following~\citep{lin2023video}.
\begin{equation}
T_{i+1} = LLM([\phi_{proj}(F^s), \phi_{T}(T_{0:i})])
\end{equation}

where $T_{0:i}$ represents the input text tokens to LLM and $T_{i+1}$ indicates the predicted next token.
$\phi_{proj}$ and $\phi_{T}$ represents the Projector and the Text Encoder, respectively.
$[\cdot, \cdot]$ denotes concatenation.

\subsection{Training}
\textbf{Training of the Temporal Sampler.}
In this stage, we only train the Temporal Sampler under the single-frame supervision.
In essence, we employed a pseudo-labeling supervision strategy.
The pseudo-labels are initialized through single-frame annotations during the training process and are online updated around the annotated frames\footnote{More details about the generation of pseudo labels can be found in the Appendix.}.
We use the generated pseudo label to supervise the predicted anomaly score, which can effectively reduce the bias of the temporal sampler towards easily confued normality.

\textbf{Instruction Tuning.}
During this stage, we take the trimmed event clips as input and do not perform Temporal Sampler because each clip has been labeled as abnormal or normal.
In this stage, we train the projector and use LoRA~\cite{hu2021lora} to fine-tune the Multi-modal LLM.
We conduct different tuning strategy and compare them in the next section.
Given the projected visual features $F_v$ and the textual input embedding $F_t$, the LLM decode them into a sequence words $\mathcal{A}$.
we follow mainstream works to use the original auto-regressive training objective.
The objective aims to maximize the likelihood of generating the ground truth answer sequence given the input features, encouraging the model to produce coherent and accurate responses based on the input features.

\section{Experiments}
\label{sec:exp}
In this section, we conduct extensive experiments to thoroughly demonstrate the capabilities of our proposed model, \textit{i.e.}, \textit{Holmes-VAD}.
\subsection{Experiment Setup}
\label{subsec:exp_setup}
\textbf{Datasets.}
We conduct  the comparative experiments on two standard VAD datasets, namely, \textbf{UCF-Crime}~\citep{ucf} and \textbf{XD-Violence}~\citep{xdviolence}. 
(1) \textbf{UCF-Crime}~\citep{ucf} comprises 1900 untrimmed videos totaling 128 hours from outdoor and indoor surveillance cameras. It encompasses 13 classes of real-world anomalies, including \textit{Abuse}, \textit{Explosion}, \textit{Fighting}, and \textit{Shooting}. In the weakly-supervised setting, there are 1610/290 videos for training/testing, with the training set consisting of 810 abnormal videos and 800 normal videos, respectively.
(2) \textbf{XD-Violence}~\citep{xdviolence} is the largest VAD benchmark, comprising 4754 videos totaling 217 hours sourced from surveillance, movies, car cameras, and games. It encompasses 6 anomaly classes: \textit{Abuse}, \textit{Car Accidents}, \textit{Explosions}, \textit{Fighting}, \textit{Riots}, and \textit{Shooting}. The training/testing video count stands at 3954/800, adhering to a weakly-supervised framework. The training set comprises 1905 abnormal videos and 2049 normal videos, respectively.

\textbf{Metrics.}
To evaluate the anomaly detection performance of the temporal sampler, we use 
the Area Under the Curve (AUC) as the main
evaluation metric for UCF-Crime following~\citep{rtfm,S3R,MSL,URDMU,zhang2024glancevad}. 
Meanwhile, the AUC of the frame-level precision-recall curve (AP) is utilized for XD-Violence.
To evaluate the quality of explanation response,
we randomly extract 86 abnormal/normal video segments from the test videos of UCF-Crime and XD-Violence, and then invite 10 users to vote on the responses of different models from 3 aspects include {Judgement Accuracy (JA)}, {Content Perception (CP)} and {Anomaly Explanatory (AE)}. Please see the Appendix for more details about the metrics.

\textbf{Implementation details.}
In our study, 
we take the ViT in LanguageBind model~\cite{zhu2023languagebind} as the Video Encoder and initialize the Multi-modal LLM  with Video-LLaVA~\cite{lin2023video}.
UR-DMU~\cite{URDMU} serves as the foundation structure for our Temporal Sampler.
To optimize the Temporal Sampler, we ramdomly sample  one frame at 16-frame intervals, and Adam optimizer with a learning rate of 1e-4 is adopted.
Note that when evaluating performance on XD-Violence and UCF-Crime, only videos in the corresponding training sets are used to train our model for fair comparisons.
For instruction tuning, we train with a batch size of 128 for 1 epoch, using the AdamW optimizer with cosine learning rate decay and a warm-up period, setting the projector's learning rate to 2e-5.
The LoRA~\cite{hu2021lora} parameters are set as: $r$=64, $\alpha$=128, and learning rate=2e-4.
The abnormal threshold $\theta$ is set to 0.8 during inference.
Experiments are conducted on 2 NVIDIA A100 GPUs.

\subsection{Main Results}
\begin{table}[t]
\centering
\caption{Comparision with state-of-the-art Video Anomaly Detection approches. We include semi-supervised (Semi.) methods, unsupervised (Un.) methods, weakly-supervised (W.) methods and some other methods. “$\ast$” represents the result reported in ~\cite{zanella2024harnessing}.
}
\label{tab:vad_comparison}
\resizebox{0.9\textwidth}{!}{
\begin{tabular}{lccc|c|c}
\toprule
\multicolumn{1}{c}{\multirow{2}{*}{\textbf{Methods}}}  & \multirow{2}{*}{\textbf{Backbone}} & \multirow{2}{*}{\textbf{Supervision}} & \multirow{2}{*}{\textbf{Explanation}} & \multicolumn{1}{c|}{XD-Violence}  & \multicolumn{1}{c}{UCF-Crime}    \\
\cmidrule{5-6} 
\multicolumn{1}{c}{} & & & &  \multicolumn{1}{c|}{AP/\%}  & \multicolumn{1}{c}{AUC/\%}  \\ \midrule
\multicolumn{6}{c}{Non-explainable VAD} \\ \midrule
Conv-AE~\cite{ConvAE} &-  &Semi. &\False  & 27.25 & 50.60  \\
GODS~\cite{GODs} &I3D & Semi. &\False & N/A & 70.46  \\
GCL~\cite{zaheer2022generative} &ResNext  &Un. &\False & N/A & 71.04   \\
DYANNET~\cite{thakare2023dyannet} &I3D  &Un. &\False &N/A  &84.50  \\
MIST~\cite{mist} &I3D & W. &\False & N/A  & 82.30  \\
Wu~\etal~\cite{xdviolence} & I3D  & W. &\False & 78.64 & 82.44   \\
RTFM~\cite{rtfm} &I3D  & W. &\False & 77.81 & 84.30   \\
MSL~\cite{MSL} &I3D  &W. &\False &78.28  & 85.30   \\
S3R~\cite{S3R} &I3D  & W. &\False &80.26  & 85.99 \\
MGFN~\cite{chen2023mgfn} &I3D & W. &\False &79.19 & 86.98  \\
UR-DMU~\cite{URDMU} &I3D & W. &\False & 81.66 & 86.97   \\ 
CLIP-TSA~\cite{cliptsa} &ViT  &W. &\False &82.19  &87.58  \\
VadCLIP~\cite{vadclip} &ViT &W. &\False &84.51  & 88.02  \\
Yang~\etal~\cite{yang2024text} &ViT &W. &\False &83.68  & 87.79 \\
Wu~\etal~\cite{wu2023open} &ViT  &Open-Vocabulary &\False &66.53 & 86.40  \\

\midrule
\multicolumn{6}{c}{Explainable Multi-modal VAD} \\ \midrule
ZS CLIP~\cite{radford2021learning}$^*$ &ViT &Training-Free &\True &17.83  &53.16  \\
ZS IMAGEBIND~\cite{girdhar2023imagebind}$^*$ &ViT &Training-Free &\True &25.36 &55.78    \\
LLAVA-1.5~\cite{liu2023improved}$^*$ &ViT &Training-Free &\True &50.26  &72.84  \\
LAVAD~\cite{zanella2024harnessing} &ViT &Training-Free &\True &62.01 &80.28  \\
\textbf{\textit{Holmes}-VAD} (Ours) &ViT & Instruction-Tuned & \True & \textbf{90.67}  & \textbf{{89.51}} \\
\bottomrule
\end{tabular}
}
\end{table}

We compare our method with state-of-the-art methods, including semi-supervised methods~\citep{ConvAE,GODs}, unsupervised methods~\citep{zaheer2022generative,thakare2023dyannet}, weakly-supervised methods~\citep{rtfm,MSL,S3R,URDMU,cliptsa,vadclip} and recently training-free method~\cite{zanella2024harnessing}.
We have indicated their backbones, supervision methods, and performance on the UCF-Crime and XD-Violence datasets, as shown in Table~\ref{tab:vad_comparison}.
Our method has an AP of 90.67\% on XD-Violence and an AUC of 89.51\% on UCF-Crime, significantly outperforming the prior state-of-the-art methods,
which demonstrates that our method can generate less biased anomaly scores.
It is worth noting that while achieving precise localization of anomalies, Holmes-VAD is also capable of providing explanations and analysis for the detected anomalies by the model, a feature unavailable in existing non-explainable VAD methods.
Although LAVAD~\cite{zanella2024harnessing} has explainability, the training-free large language model lacks an understanding of anomaly knowledge due to the limitation of insufficient supervised data.

\subsection{Analytic Results}
\input{tables/analytic_results}

\textbf{Influence of varied training strategies on anomaly explanation.} We conduct a user study to evaluate three different training strategies over 86 test samples and 10 volunteers: 
a) Training-free: no fine-tuning;
b) Projector: fine-tuning on VAD-Instruct50k, only training the projector while keeping the Multi-modal LLM fixed; 
c) Projector+LoRA: fine-tuning on VAD-Instruct50k, training the projector and using LoRA~\cite{hu2021lora} to fine-tune the Multi-modal LLM.
As shown in Table~\ref{tab:user_study}, Projector+LoRA provide the most detailed response (46.13 words in average) and reaches the highest Judgement Accuracy (86.0\%). Addtionally, it also achieves the highest voting rate, including 61.2\% on Content Perception and 51.9\% on Anomaly Explanatory, these demonstrate better interpretability by fine-tuning Multi-modal LLM on VAD-Instruct50k.

\textbf{Backbone and supervision matters in Temporal Sampler.}
In Table~\ref{tab:ablation_backbone_supervision}, we ablate the impact of video backbone and the supervision for Temporal Sampler.
We use UR-DMU~\cite{URDMU} as our baseline method.
The results indicate that on XD-Violence dataset, LanguageBind~\cite{zhu2023languagebind} as a backbone outperforms I3D~\cite{i3d} significantly, whereas the opposite is observed on UCF-Crime.
Additionally, single-frame supervision significantly enhances performance regardless of the backbone used, demonstrating the effectiveness of point supervision in improving anomaly localization capabilities.

\noindent\textbf{Influence of perturbed single-frame annotations.}
To assess the robustness of our method to the perturbed temporal position of single-frame annotation, we introduce varied temporal timestamp shifts to the original positions of the annotated frames. As shown in Table~\ref{tab:ablation_point}, there is no significant performance degradation of our model under perturbed annotation positions, indicating that our method possesses a notable tolerance towards variations in degraded supervision.

\textbf{Temporal Sampler v.s. Uniform Sampler.}
We replace the Temporal Sampler with Uniform Sampler while maintaining the frame rate.
The video is then divided into non-overlapping clips, which are sequentially fed into the Multimodal LLM to output results. 
If the output is "Yes" the anomaly scores of all frames in the input segment are set to 1, otherwise, they are set to 0.
Finally, we compare the detection performance and inference efficiency in Table~\ref{tab:sample_method}. 
The results demonstrate that the Temporal Sampler ensures higher inference efficiency while maintaining accurate detection results.

\begin{figure*}[t]
\centering
\includegraphics[width=\textwidth]{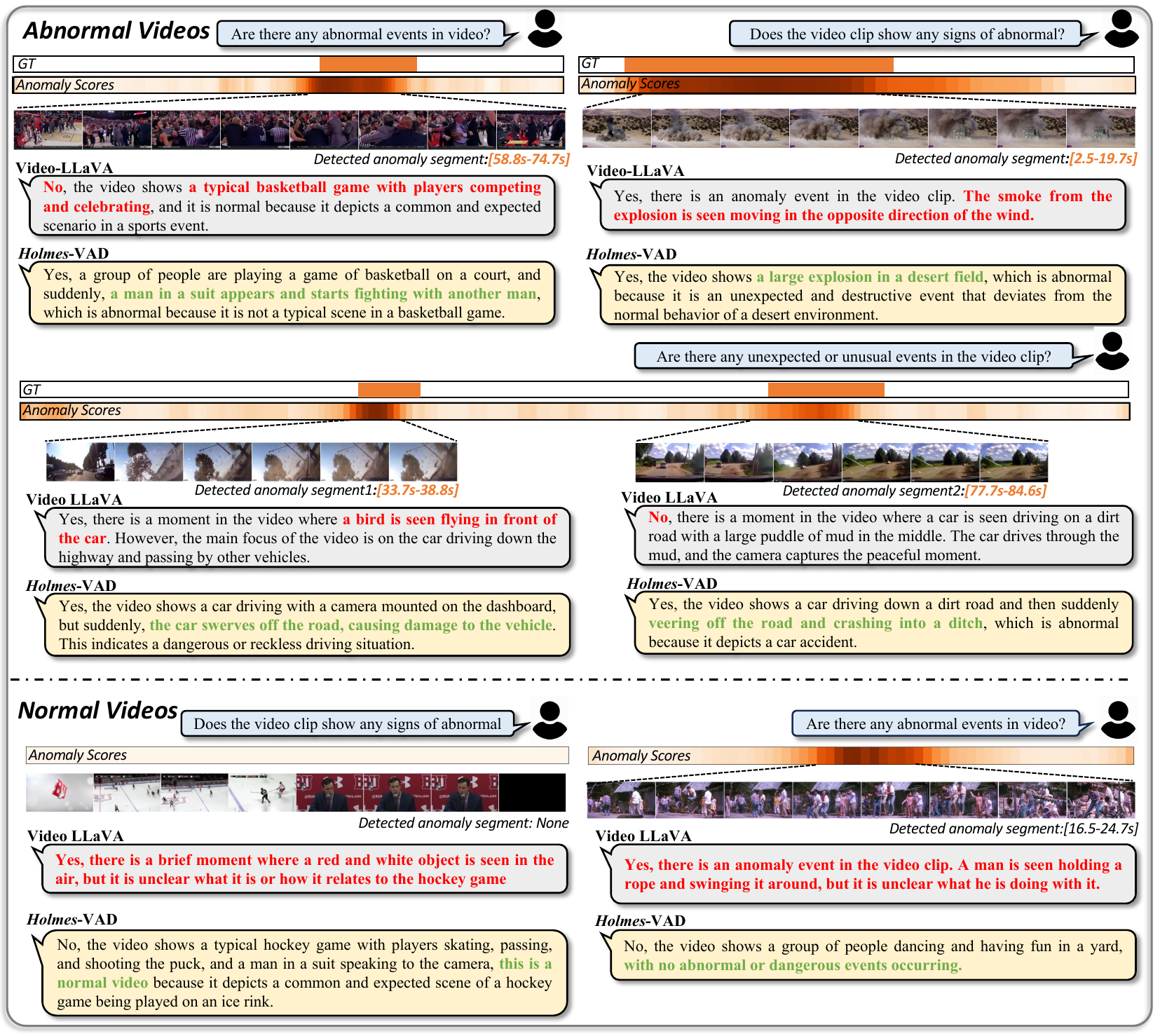}
\vspace{-4mm}
\caption{\textbf{Qualitative results.} 
We compare our interpretability results with Video-LLaVA~\cite{lv2024video} (without instruction tuning).
Correct and wrong explanations are
in \textcolor[rgb]{0,0.6875,0.3125}{green} and \textcolor{red}{red}, respectively.}
\vspace{6mm}
\label{fig:qualitative_results}
\end{figure*}
\textbf{Qualitative comparision.} To provide a more intuitive understanding of the capabilities of MLLM in explaining complex anomalies, we provide qualitative comparisons between Holmes-VAD and Video-LLaVA in Fig. \ref{fig:qualitative_results}. The results demonstrate that Holmes-VAD can accurately identify anomalies in videos and provide specific explanations for conflicts in sports competitions, explosions, and accidents captured by car cameras (Abnormal Cases). Even for normal videos, Holmes-VAD exhibits robust analytical abilities, correcting erroneous responses from the Temporal Sampler (Normal Cases). These findings highlight the effectiveness and advantage of Holmes-VAD in perceiving video events and analyzing anomalies.

\section{Conclusion}
In this paper, we introduce a video anomaly detection system called Holmes-VAD to address the biases and lack of interpretability in existing anomaly detection methods.
By introducing a more efficient labeling paradigm and constructing a large-scale multimodal video anomaly detection dataset, VAD-Instruct50k, we validated the generality and interpretability of Holmes-VAD.
Through extensive experiments, we positioned Holmes-VAD as a valuable tool for real-world applications.

\noindent\textbf{Limitation and Future work.} 
\label{subsec:limitation_futurework}
Despite the human effort for filtering the noise instruction data during constructing the VAD-Instruct50k dataset,
the reliance on off-the-shelf video captioning models for generating video description may not always capture the nuances and context-specific information.
This is a trade-off we have made between labeling costs and efficiency, we believe that the quality of data is no less important than the quantity of data, and we plan to further enhance data quality and quantity within acceptable labor costs in the future.
Furthermore, although we control the length of the video input to the Multi-modal LLM through Temporal Sampler and accurately analyze abnormal content in the trimmed video clips, there is still a lack of an effective solution for Multimodal LLM to understand long-term video anomalies without compromising its image-level perceptual capabilities.
We leave these for our future exploration.

\noindent\textbf{Acknowledgement}
This work is supported by the National Natural Science Foundation of China under grant U22B2053 and 623B2039.

{
\small
\bibliographystyle{plainnat}
\bibliography{main}
}
\newpage
\appendix
\section{Appendix}
\subsection{Broader Impact}
\label{subsec:broader_impact}
The paper proposes a video anomaly detection framework, namely \textbf{\textit{Holmes-VAD}},  that is capable of temporally identifying anomalies accurately and providing insightful explainations across even hour-long videos.
Additionally, this paper provides \textbf{VAD-Intruct50k}, a large scale multimodal video anomaly detection datasets, including single-frame annotations for untrimmed videos, and a large amount of instruction conversation data for trimmed abnormal/normal video clips.

The positive societal impacts of the work include:
\begin{itemize}
\item \textbf{Improved public safety}: The development of more accurate and interpretable video anomaly detection systems can enhance public safety by enabling quicker and more precise identification of anomalies in surveillance videos, such as criminal activities or accidents.

\item \textbf{Advancement in supervised and open-world VAD research}: The proposed VAD-Intruct50k dataset provide a). accurate temporal timestamp of the abnormal events in videos, and b). video-explanation pair for both abnormal and normal video clips,  which can pave the way for further supervised and open-world research in the video anomaly detection area.
\end{itemize}

The negative societal impacts may include:
\begin{itemize}
\item \textbf{Privacy concerns}: The use of video surveillance technology, especially in public spaces, raises concerns about privacy and the potential for intrusive monitoring of individuals without their consent.

\item \textbf{Disregard for minor anomalies}: Despite efforts to reduce bias in anomaly detection, there is still a risk of disregard for subtle anomalies such as stealing in the supermarket, leading to potential undetected anomalies.
\end{itemize}
Consequently, researchers should adhere to relevent laws and regulations, and strive to avoid using our model or dataset for any improper invasion of privacy. Meanwhile, all our model and data will be only used for research purpose to avoid the potential negative societal impacts.

\subsection{Process of single-frame annotation.}
\label{supp:single_frame}
\begin{figure}[h]
\centering
\includegraphics[scale=0.4]{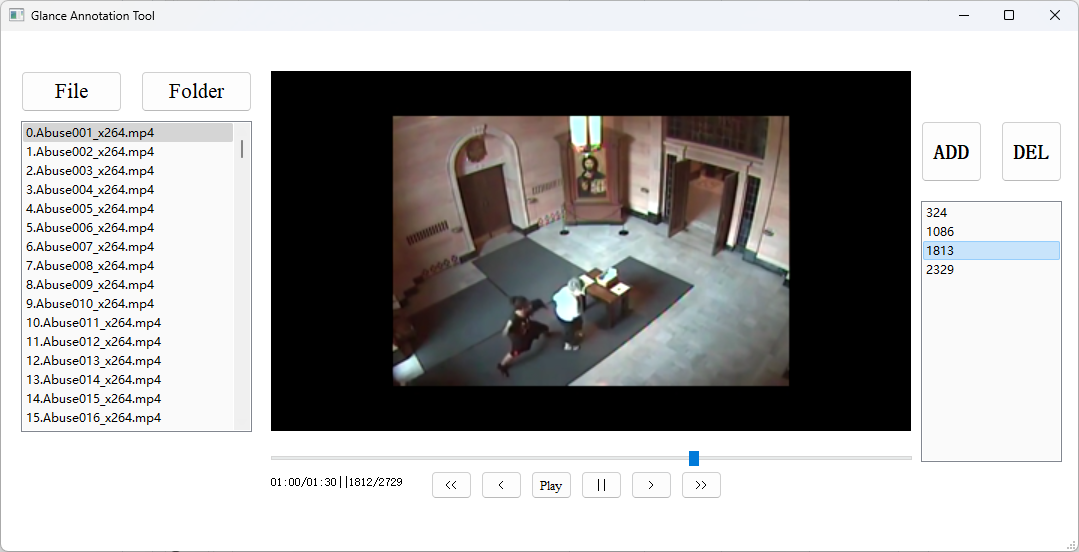}
\caption{Screenshot of the single-frame annotation interface.
}
\label{interface}
\end{figure}
\textbf{Annotation tool.}
We develop an interface designed specifically for single-frame annotation in videos, as shown in Fig.~\ref{interface}.
This interface makes it easier to navigate through video lists, adjust video progress rapidly, and automatically record timestamps for annotating individual frames.
Furthermore, it enables the preview of annotated frames. By clicking on the annotated frame ID, the video progress automatically synchronizes with the corresponding temporal position.
These features greatly streamline the annotation process, enhancing convenience and efficiency.
If annotators come across any errors or need to make adjustments, they can delete incorrect annotations and proceed with re-annotation.

\textbf{Quality control.}
We initially divide the entire dataset into various portions and distribute them among different annotators for labeling. Once the first round of annotations is completed, we proceed with a secondary review of the video annotations to eliminate incorrect or redundant annotations.
In addition, we include ignored clicks to minimize the possibility of overlooking potential anomalies. This process ensures the \textbf{Reliability} and \textbf{Comprehensiveness} of the single-frame annotations.

\begin{figure}[t]
\centering
\includegraphics[scale=0.32]{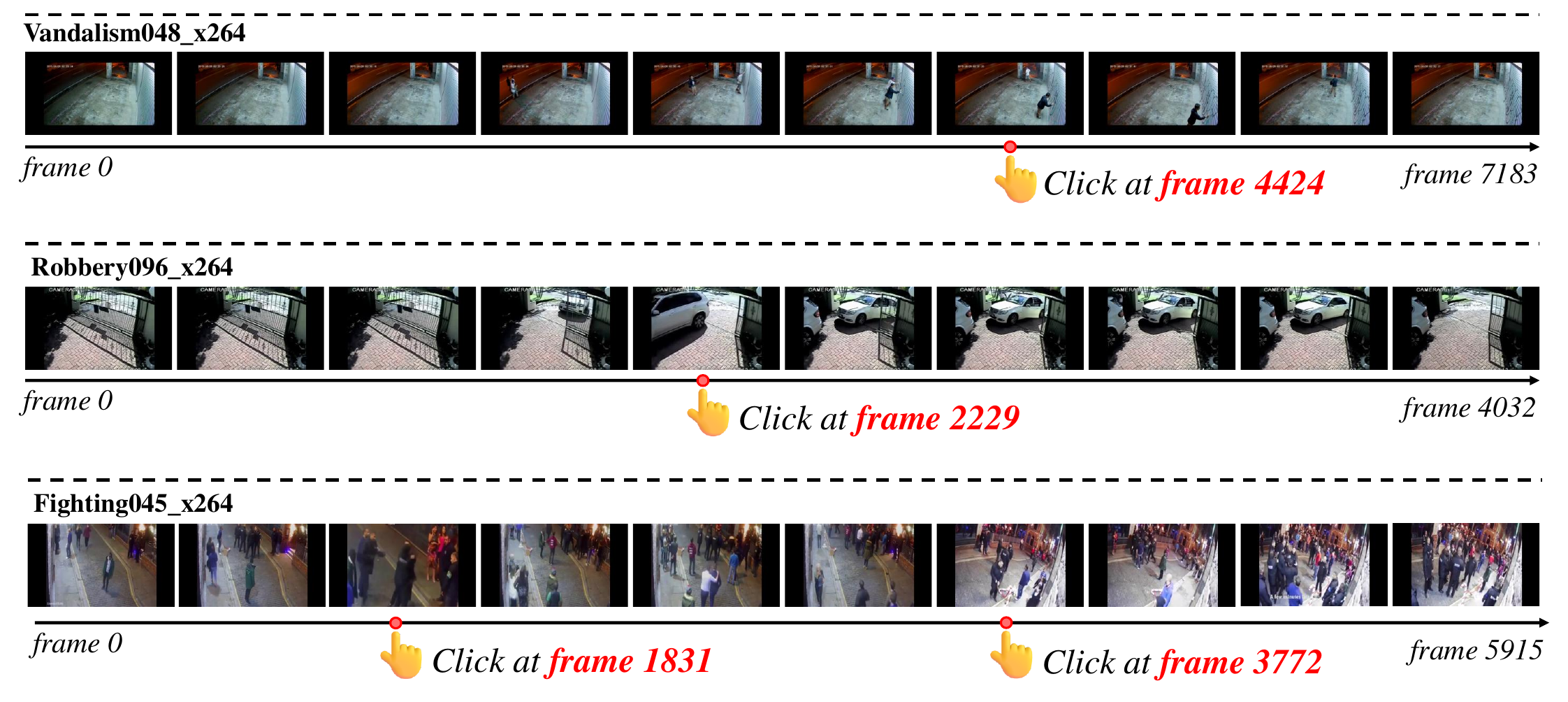}
\caption{Examples of single-frame annotation.}
\label{glance_examples}
\end{figure}

\textbf{Examples of single-frame annotation.}
To facilitate a better understanding of the annotation process, we offer several examples of annotated videos in Fig.~\ref{glance_examples}.

\subsection{Model architecture and training details of the Temporal Sampler.}
\label{supp:temporal_sampler}
\begin{wrapfigure}{r}{0.35\textwidth}
  \centering
  \vspace{-4mm}
  \includegraphics[width=0.25\textwidth]{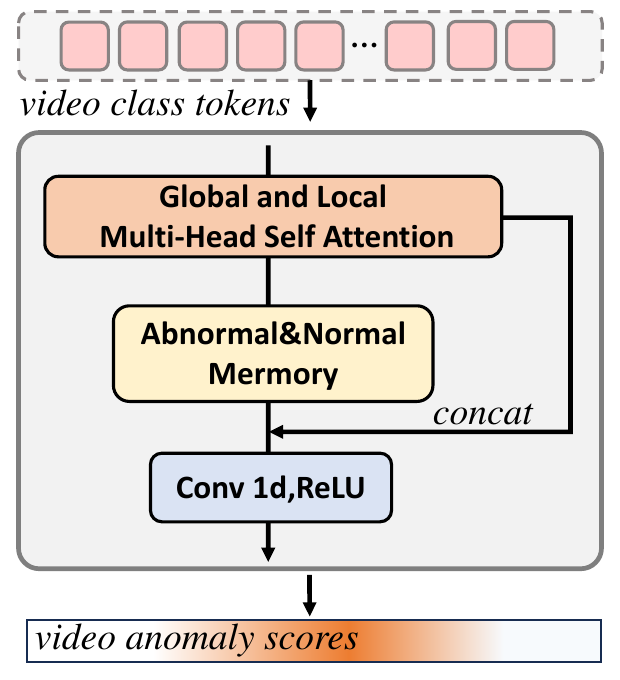}
  \caption{Architecture of the Temporal Sampler.}
\label{baseline_methods}
\end{wrapfigure}
\textbf{Model architecture.}
We use UR-DMU~\cite{URDMU} as the VAD network in our Temporal Sampler.
As shown in Fig.~\ref{baseline_methods}, UR-DMU utilizes a Global and Local Multi-Head Self Attention (GL-MHSA) module to capture both long-range and short-range temporal relationships among video snippets. Furthermore, UR-DMU introduces two memory banks to store and differentiate abnormal and normal prototypes, thereby maximizing the margins between these two representations. In order to learn discriminative representations, UR-DMU employs triplet loss to increase the feature distance after interacting with different memories. Simultaneously, it utilizes KL loss to constrain the normal memory to follow a Gaussian distribution, accounting for the variance introduced by noise. Thus, the base loss function for the UR-DMU baseline is defined as follows:
\begin{equation}
{\mathcal L}_{base} = {\mathcal L}_{mil} + {\mathcal L}_{mag} + {\mathcal L}_{triplet} + {\mathcal L}_{kl}
\end{equation}
\textbf{Training details.}
During the training stage of the Temporal Sampler, we leverage the sparse single-frame annotations to generate reliable dense snippet-level pseudo label.
As illustrated in Alg.~\ref{alg:alm}, we employ a dynamic threshold and perform local bidirectional mining based on the single-frame annotations.
Snippets with anomaly scores exceeding a specific proportion of the annotated snippet's score are identified as pseudo anomaly snippets. We set $\alpha=0.9$ in our implementation.
\begin{algorithm}[h]
\caption{Pseudo Label Mining.}
\label{alg:alm}
\raggedright
\textbf{Input}: Anomaly score $\mathbf{S}\in{\mathbb R}^T$, single-frame annotations $\mathcal{G}=\{g_i\}^{N_g}$, anomaly ratio $\alpha$.\\
\textbf{Output}: Pseudo anomaly snippets $\mathcal{T}^a=\{t_i\}_i^{N_{a}}$.

\begin{algorithmic}[1] 
\STATE Let $\mathcal{T}^a \gets \varnothing$.
\FOR{every $g_i \in \mathbf{G}$}

    \FOR{$t = g_i$ {\bfseries to} $g_{i-1}$}
        \STATE \textbf{if} $\mathbf{S}[t] > \alpha \cdot S[g_i] $, \textbf{then} $\mathcal{T}^a \gets t \cup \mathcal{T}^a$, \textbf{else break}
        \STATE \textbf{end if}
    \ENDFOR

    \FOR{$t = g_i$ {\bfseries to} $g_{i+1}$}
        \STATE \textbf{if} $\mathbf{S}[t] > \alpha \cdot S[g_i]$, \textbf{then} $\mathcal{T}^a \gets t \cup \mathcal{T}^a$, \textbf{else break}
        \STATE \textbf{end if}
    \ENDFOR
\ENDFOR
\STATE \textbf{Return} $ \mathcal{T}^a$
\end{algorithmic}
\end{algorithm}
After mining the pseudo anomaly snippets, we adopt Gaussian function to smooth the binary pseudo label:
\begin{equation}
\hat{S}(\mathbf{t}) = \textbf{norm}(\sum_{i=1}^{N_a}{exp(-\frac{\|\mathbf{t}-\mathbf{t_i}\|^2}{2r^2})})
\label{eq:splatting}
\end{equation}
where $r=0.1$ indicates the smoothing ratio.
We use the generated dense and smooth pseudo label to supervise the predicted anomaly score:
\begin{equation}
{\mathcal L}_{abn} = BCE(S,\hat{S})
\end{equation}
where $S$ and $\hat{S}$ denote the predicted anomaly score and the generated pseudo frame-level label, respectively.

\subsection{Details of human evaluation.}
\begin{figure}[h]
\centering
\includegraphics[scale=0.3]{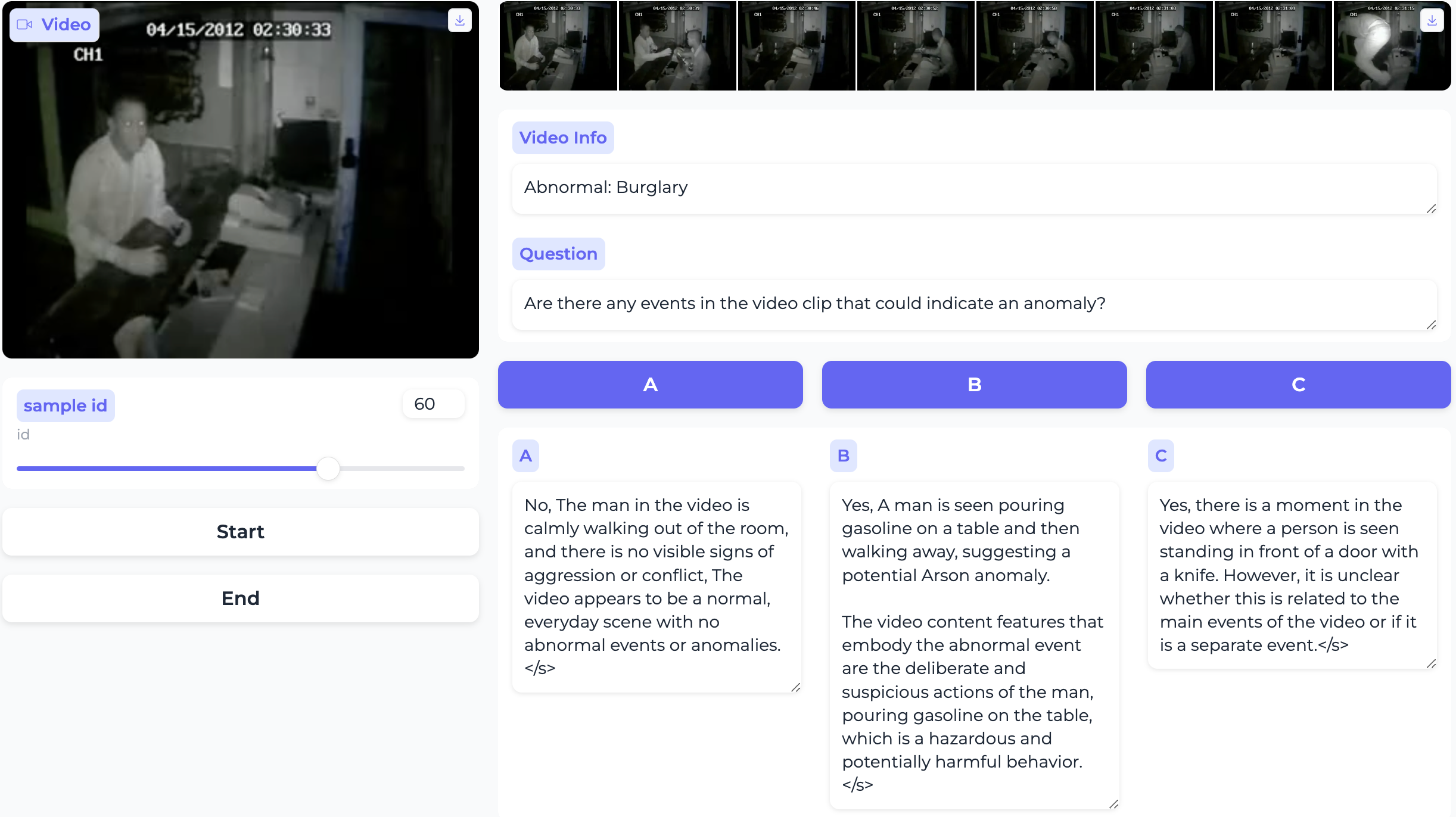}
\caption{Screenshot of the human evaluation interface.}
\label{human_eval_interface}
\end{figure}
\clearpage
\begin{figure}[h]
\centering
\includegraphics[width=\textwidth]{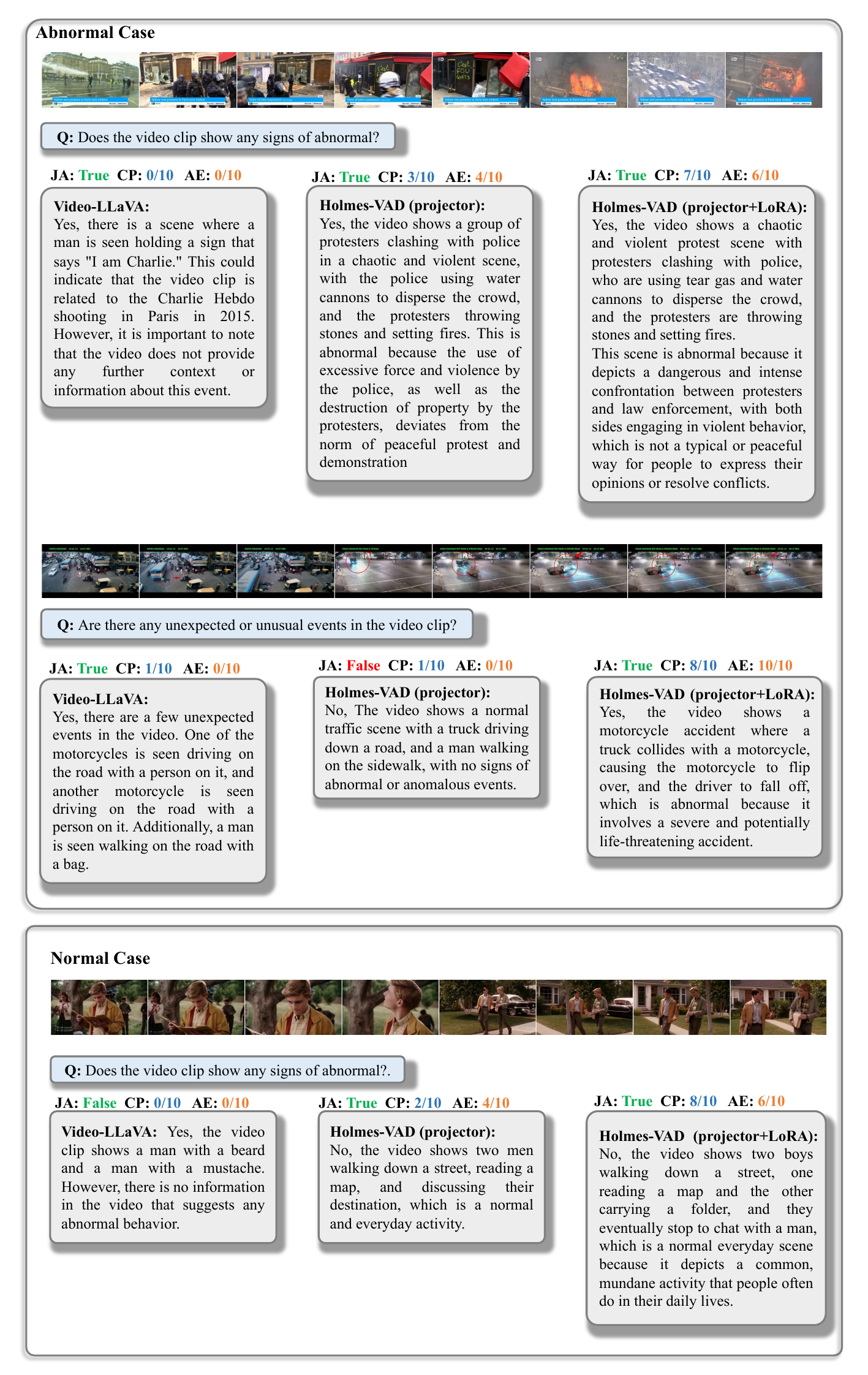}
\vspace{-4mm}
\caption{\textbf{Qualitative comparision in human evaluation.} We show the results of Judgement Accuracy (JA), Content Perception (CP) and Anomaly Explanatory (AE) above the answer box of each model.}
\label{fig:qualitative_supply}
\end{figure}
To evaluate the quality of explanation response,
we randomly extract 86 abnormal/normal video segments from the test videos of UCF-Crime and XD-Violence, and then invite 10 users to vote on the responses of different models from 3 aspects include {Judgement Accuracy (JA)}, {Content Perception (CP)} and {Anomaly Explanatory (AE)}.
\begin{itemize}
    \item \textbf{Judgement Accuracy (JA)}: Determine whether the model's judgment on anomalies is correct, we extract predictions by matching "Yes"/"No" in the answers, and compare them with the ground truth labels (abnormal/normal). Finally, we calculate the accuracy of the judgments.
    \item \textbf{Content Perception (CP)}: The accuracy and clarity of the model's descriptions of the content, characters, and events in the video scenes, as well as any potential hallucination issues (descriptions of non-existent objects in the video or responses unrelated to the questions).
    \item \textbf{Anomaly Explanatory (AE)}: The model's ability to analyze and interpret abnormal/normal events in the video.
\end{itemize}
We provide the screenshot of the human evaluation interface in Fig.~\ref{human_eval_interface}, to ensure a fair selection, the names of the models are not visible to the users, and choices can only be made from anonymous options.
In Fig.~\ref{fig:qualitative_supply}, we provide several test examples, with the results of the Judgement Accuracy (JA), Content Perception (CP) and Anomaly Explanatory (AE).

\subsection{Dats statistical analysis of VAD-Instruct50k}

In Table~\ref{tab:dataset_statistic}, we conduct a statistical analysis of our proposed \textbf{VAD-Instruct50k} and compare it with representative datasets in the VAD field, which  shows the significant volume and excellent diversity of our constructed instruction dataset.
\begin{table}[h]
\centering
\caption{\textbf{Datasets Statistics}.}
\label{tab:dataset_statistic}
\setlength{\tabcolsep}{4pt}
\begin{tabular}{l|cccc}
\toprule
Dataset & \#Videos & Annotation Type  & \#Queries & Avg word  \\
\midrule
CHUK Avenue &37 &None &N/A &N/A \\
ShanghaiTech &437 &None/video label &N/A &N/A \\
UCF-Crime~\cite{ucf} & 1,610 & video label &N/A & N/A  \\
XD-Violence~\cite{xdviolence} & 4,754 & video label &N/A & N/A \\
UCA~\cite{yuan2023surveillance} & 1,854 & segment caption & 23,542 & 20.15  \\
\midrule
\textbf{VAD-Instruct50k (Ours)}  & \textbf{5,547} & \textbf{single-frame\&segment instruction} & \textbf{51,567} & \textbf{44.83}  \\
\bottomrule
\end{tabular}

\end{table}

\end{document}